# Multimodal Deep Learning for Scientific Imaging Interpretation

*Abdulelah S. Alshehri†‡, Franklin L. Lee†, Shihu Wang†.*

†Science and Technology Division, Corning Incorporated, Corning, NY 14831, USA
*{Alshehria, LeeFL, WangS7}@corning.com*
‡Robert Frederick Smith School of Chemical and Biomolecular Engineering, Cornell University, Ithaca, NY 14853, USA
*Asa279@cornell.edu*

**Abstract**

In the domain of scientific imaging, interpreting visual data often demands an intricate combination of human expertise and deep comprehension of the subject materials. This study presents a novel methodology to linguistically emulate and subsequently evaluate human-like interactions with Scanning Electron Microscopy (SEM) images, specifically of glass materials. Leveraging a multimodal deep learning framework, our approach distills insights from both textual and visual data harvested from peer-reviewed articles, further augmented by the capabilities of GPT-4 for refined data synthesis and evaluation. Despite inherent challenges—such as nuanced interpretations and the limited availability of specialized datasets—our model (GlassLLaVA) excels in crafting accurate interpretations, identifying key features, and detecting defects in previously unseen SEM images. Moreover, we introduce versatile evaluation metrics, suitable for an array of scientific imaging applications, which allows for benchmarking against research-grounded answers. Benefiting from the robustness of contemporary Large Language Models, our model adeptly aligns with insights from research papers. This advancement not only underscores considerable progress in bridging the gap between human and machine interpretation in scientific imaging, but also hints at expansive avenues for future research and broader application.

# Table of Contents

# Introduction

In the intricate landscape of materials science, analyzing the relationships between a material's diverse attributes—such as morphology, phase, atomic configurations, and interfacial structures—and its physical or chemical properties and behaviors is a central task. This analysis paves the way for fine-tuning synthesis parameters towards the creation of materials tailored for diverse applications, from pharmaceuticals and catalysts to battery materials (Ge, et al. 2020). Instrumental to this design process is the reliance on advanced characterization techniques. Particularly, microscopic imaging techniques, such as Scanning Electron Microscopy (SEM), which stand out for their capacity to provide unparalleled real-space visualizations of material structures across scales. Nevertheless, while these scientific imaging modalities have revolutionized material insights, the divergence of their imagery from routine human visual experiences can lead to variability in interpretations. Crucially, their interpretation can be influenced by a wide range of human factors, with the expertise of the observer being particularly significant (Goldstein, et al. 2017).

Deep learning, a notable subset of machine learning, has catalyzed unprecedented advancements in materials science, delivering advancements that eclipse years of conventional experimentation and theorization (Alshehri and You 2022). Its feats are evident in areas such as protein folding (Jumper, et al. 2021), synthesis planning (Coley, et al. 2017), and molecular design (Yao, et al. 2021). While traditionally deep learning has outperformed conventional AI algorithms and human performance in intricate tasks, including computer vision (CV) and natural language processing (NLP), these applications often circumscribe their predictive capacity to predefined outputs, be it a class in classification tasks, a quantified outcome in regression, or a structured output in design and generative paradigms. A conspicuous void in the broad adoption of deep learning within materials science lies in integrating the nuanced reasoning embedded within the linguistic folds of research literature, where measurements and classifications are woven with linguistic descriptions.

The rise of large language models (LLM) signifies a transformative shift across various domains. Expanding beyond traditional NLP tasks, these models now act as platforms enabling sophisticated, human-assistant-like exchanges, notably venturing into the chemical space (Vert 2023). As LLMs make inroads into materials science, fine-tuned LLMs have occasionally outperformed benchmarks set by GPT-4 (OpenAI, GPT-4 Technical Report 2023), offering enhanced contextual discovery and optimization (Jablonka, et al. 2023). While their primary strength appears to be assisting chemical experts and democratizing access for novices, a significant impediment to their full-fledged integration in design and optimization remains (Bran, et al. 2023). This challenge stems from their overwhelming reliance on textual data, which can inadvertently neglect critical facets, such as the three-dimensional conformational interactions of drug-like molecules, commonly abstracted into two-dimensional representations as SMILES (Weininger 1988).

Emerging in this context is LLM-powered multimodal deep learning, poised to revolutionize materials science research. This approach can integrate diverse data forms—including visual, graph-based, and textual (OpenAI, GPT-4 Technical Report 2023) (Akkus, et al. 2023). Despite challenges in data representation and integration, these models adeptly combine visual inputs (e.g., SEM images) with textual information from research literature. By harmonizing these modalities, they leverage the strengths of each, providing in-depth insights into materials' properties and behaviors. This intersection of human expertise and advanced AI augments the trajectory of materials research.

In this study, we delve into the potential of multimodal deep learning within materials science, focusing on the interplay between visual and textual information. We assess the ability of current question-answering multimodal LLMs in interpreting SEM images of glass materials under varying levels of question complexity and context. Further, we introduce rigorous evaluation metrics to objectively measure response quality, probe the context-driven descriptions based on SEM images, and scrutinize the model's capacity for identifying unseen features. First, we detail our methodology, encompassing data collection, processing, deep learning model, and assessment methods. Subsequently, we present our findings, elucidating and discussing key insights and potential limitations, respectively. We conclude with a summary and contemplation on the broader ramifications and prospective avenues for embedding multimodal deep learning in the realm of materials science.

# Methods

The methodology employed in this study is a comprehensive amalgamation of data extraction, generation, and processing, followed by the implementation of a deep learning model and rigorous evaluation methods. This section provides a detailed account of the various steps involved in our research, from the initial stages of data collection to the final evaluation of the model's performance. We delve into the specifics of our data extraction and generation process, highlighting the importance of a curated dataset and the transformative capabilities of GPT-4 in generating precise question-answer pairs. We then discuss the deep learning model employed, elucidating its architectural details and the process of fine-tuning it for our specific application. Finally, we introduce our evaluation methods, which encompass overall quality assessment, context assessment, and feature identification assessment, providing a comprehensive framework for assessing the machine interpretation of scientific imaging.

## Data Extraction and Generation

The process of data extraction and generation forms the bedrock of our work, providing the necessary input for our deep learning model. This subsection outlines the meticulous process of curating a dataset from glass materials research, ensuring focused model training and representative sampling within a specific domain. We detail the systematic extraction of research papers, SEM images, and their corresponding textual data, highlighting the pivotal role of a human oracle in validating and rectifying any data irregularities. Further, we delve into the generation of a diverse question-answer dataset, leveraging the capabilities of GPT-4. This process involves the classification of questions based on their overarching theme and depth, culminating in a comprehensive dataset ready for model training. The subsequent paragraphs provide a granular description of our data extraction and generation process, elucidating the details involved in each step.

Data Extraction

In our endeavor to leverage multimodal deep learning in materials science, we curated a dataset primarily from glass materials research, aiming for focused model training and representative sampling within a specific domain. This approach ensures the model's capability to generalize the intricacies of descriptions intrinsic to glass materials without extensive data collection. For this purpose, we assembled a dataset of 72 research papers published post-2010 to acquire images with consistent resolution and uncompromised quality. Predominantly, our collection comprises full research articles; however, it also encompasses a limited number of shorter communications and conference papers. These papers were systematically extracted using the keywords "glass" and "scanning electron microscopy" from three publishers: American Chemical Society, Elsevier, and Springer. From the acquired papers, we extracted SEM images, their references (for example, 'Fig. Xa,' where 'X' represents the figure number as it appears in the main text of the extracted paper), main text, and DOIs. This involved process incorporated manual techniques, supported by automated image processing and PDF extraction libraries. Throughout this extraction process, a human oracle played a pivotal role by ensuring validation and rectification of any data irregularities or discrepancies encountered. With the refined dataset in hand, 62 papers (encapsulating 404 SEM images) were earmarked for training, while the remaining 10 papers (and their corresponding 77 SEM images) were set aside for evaluation, aiming for an approximate 85%-15% split across papers and images. This rich curated data was then fed to GPT-4, facilitating the generation of a diverse question-answer dataset, the intricacies of which are expounded in the subsequent subsection. For an immediate reference, Table S1 in the supporting information furnishes a comprehensive list of figure names and the respective extracted paper DOIs. We also apply clustering analysis that groups the training and evaluation papers in separate clusters, provided in Figure S1 and its description.

Data Generation

Leveraging the transformative capabilities of advanced language models such as GPT-4 (OpenAI, GPT-4 Technical Report 2023), we explored the uncharted territory of multimodal deep learning in generating precise question-answer pairs, streamlining the process, and reducing the necessity of expensive human involvements. Our methodology consisted of introducing the full text, image citations, and a detailed set of guidelines into the GPT-4 system. The model was then optimized to classify questions based on two critical parameters: (1) the overarching theme termed as 'category' and (2) the depth or intricacy of the question denoted as 'subcategory'. This rigorous approach culminated in the generation of 4291 question-answer pairs for the training set and an additional 757 for evaluation. A granular description, divided by category, is provided in Table 1. It is worth noting that the generated 'questions' are not exclusively phrased as questions; they can also be framed as prompts, such as 'describe,' 'identify,' and so forth.

*Table 1: The breakdown of training and evaluation data per category. The percentages represent the composition of each category within the respective training (middle column) and evaluation (right column) sets.*

| Category | Percentage (Training: Evaluation) | |
|---|---|---|
| General | 18.5% | 19.0% |
| Microstructure | 8.2% | 12.9% |
| Morphology | 12.8% | 16.9% |
| Phase and Composition | 14.1% | 11.9% |
| Defect and Fracture | 12.0% | 14.9% |
| Processing and Environmental Impact | 17.0% | 9.2% |
| Prediction and Interpretation | 17.4% | 15.1% |
| **Total (Count)** | **4291** | **757** |

To clarify our categorization approach, we provide its structure, emphasizing its relevance to data typically extracted from SEM images. Beyond a simple taxonomy, this framework can also serve as a general tool, adaptable for crafting intricate queries suited to scientific images and accompanying literature descriptions. Table 2 below provides an overview of our categorization approach. For each category, we include a description and sample questions or prompts that guide the analysis of SEM images and accompanying descriptions.

*Table 2: Categorization framework for SEM images, listing categories, their descriptions, and sample questions or prompts.*

| Category | Description | Sample Questions/Prompts |
|---|---|---|
| **General** | This category encompasses fundamental facets of the SEM image, such as material identification, image scaling, and salient feature interpretation | Describe the materials depicted in the image. |
| | | Is the sample homogeneous or heterogeneous? |
| **Microstructure** | This category delves into the material's microstructure, highlighting elements like grain structures and boundaries. | Describe the shown microstructure in the image. |
| | | Are grain boundaries visible and, if so, how are they characterized? |
| **Morphology** | Detailing microscopic physical attributes, this segment covers particle shapes, sizes, arrangements, and surface topography. | How would one describe the figure's morphology? |
| | | Can you detail the particles' sizes, shapes, and arrangements? |
| **Phase and Composition** | Addressing the chemical make-up, this category highlights various phases, elements, and potential impurities. | What phases are shown in the figure? |
| | | Can the sample's chemical composition be detailed, considering diverse elements and impurities? |
| **Defect and Fracture** | Focused on material irregularities and breakage patterns, this category addresses imperfections like voids and cracks | Are any material defects or anomalies evident? If present, how would they be detailed? |
| **Processing and Environmental Impact** | Assessing factors like sample creation and manufacturing techniques, this category evaluates environmental ramifications. | How was the sample prepared and processed? |
| | | What are the potential environmental impacts of these processes? |
| **Prediction and Interpretation** | Leveraging insights from prior categories, this section foresees material behavior and evaluates any further interpretations. | Describe any predictions that can be made about the behavior or the performance of the sample in the figure. |
| | | How is the material's expected longevity interpreted from the image? |

By organizing the SEM image analysis into these categories, our approach provides a systematic and comprehensive method for extracting and understanding information. On the subcategory level, we further classify questions as simple or complex for all categories, except for the general category, which only includes simple questions. The simple subcategory necessitates basic facts about the SEM sample and basic analysis, while the complex subcategory demands a deeper understanding, analysis, and interpretation of the SEM sample.

To ensure the generation of high-quality question-answer pairs, we employed a set of rigorous rules, which are detailed in the Supporting Information as Prompt S1. Some of the most important rules include: (1) generating pairs strictly based on the paper's description, details, and analysis of a specific SEM image; (2) ensuring that questions and answers are unique, diverse, factually and grammatically correct; (3) avoiding the use of abbreviations or paper-specific notations in the pairs; and (4) providing answers that are more than two sentences long, rich in relevant paper information, and detailed descriptions. These rules were followed to maintain the integrity and relevance of the generated question-answer pairs, ultimately contributing to the robustness of our multimodal deep learning model.

## Deep Learning Model

The past few years have seen remarkable progress in the development of vision-language models, which are trained on pairs of images and text (Radford, et al. 2021, Wang, et al. 2021, Yuan, et al. 2021). These models have been primarily utilized for specific downstream tasks, such as image captioning, image-text retrieval, and visual question answering (Goyal, et al. 2017, Wang, et al. 2022). However, our focus is on the domain of instruction-following, where models typically have operated within agent-like environments or systems, coordinating multiple deep learning models for task execution. Models such as the image-editing model InstructPix2Pix (Brooks, Holynski and Efros 2023) and cross-modal pretrained models like X-GPT (Xia, et al. 2021) exemplify these approaches. Despite the similarities in techniques and nature, our approach differs by employing an end-to-end multimodal deep learning model for multiple tasks.

Specifically, we extend the Large Language and Vision Assistant Model (LLaVA) (Liu, et al. 2023) to scientific imaging applications.

The LLaVA architecture represents a pioneering extension of instruction tuning LLMs (Liu, et al. 2023) into the multimodal space, effectively leveraging the capabilities of LLMs and vision models. In the NLP community, the LLM instruction-tuning technique has been pivotal in enabling LLMs such as GPT-4 (OpenAI, GPT-4 Technical Report 2023) to follow natural language instructions and complete real-world tasks, leading to the development of instruction-tuned counterparts like ChatGPT (OpenAI, ChatGPT 2023). Inspired by these advancements, the LLaVA approach seeks to enhance the LLM model's instruction-following abilities when presented with visual data. This is achieved through the end-to-end training of an LLM connected to a vision encoder, resulting in state-of-the-art performance on Science QA (Lu, et al. 2022), a multimodal reasoning dataset.

Building upon this progress, we extend the LLaVA model to scientific imaging with minor modifications. While we retain all architectural details, including the vision encoder as CLIP ViT-L/14, we opt for the 7B OpenLLaMA (Geng and Liu n.d.) reproduction of LLaMA (Touvron, et al. 2023) due to the latter's unavailability. For an input SEM image, the vision encoder outputs a trainable projection of visual features. The projection is combined with the language instruction or "question" embeddings in a simple linear layer that consolidates the input to the OpenLLaMA model, which ultimately generates the LLM response or answer. Figure 1 provides a high-level overview of GlassLLaVA architecture.

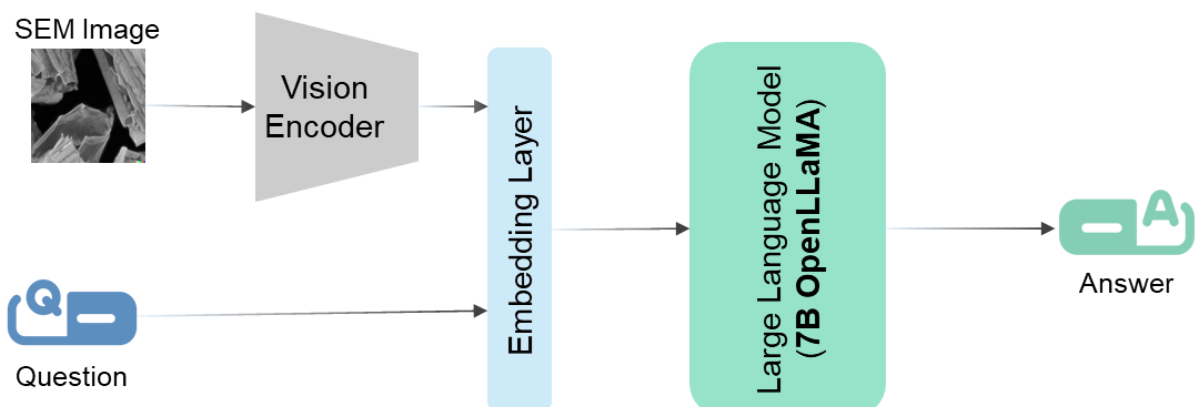

*Figure 1: High-level description of the GlassLLaVA architecture.*

Our model, GlassLLaVA, undergoes a two-step process of "pretraining" and "fine-tuning" constructed on the LLaVA COCO dataset, as detailed in the paper and documentation (Liu, et al. 2023). Subsequently, we fine-tune our model on our datasets using 8 Nodes, each comprising an NVIDIA A100 GPU (40GB memory), 32 CPUs, and 8GB RAM. The training process spans approximately 3 hours for 28 epochs, with a batch number of 4 and a learning rate of 1e-5, employing a cosine scheduler that reduces the learning rate at each step. Although we did not observe a change in the model loss after the 12th epoch at 0.003, we noted slight improvements in the quality of answers when comparing models trained at the same starting learning rate and beyond 12 epochs. However, it is worth noting that we did not quantify this difference but rather assessed the quality of answers to a small subset of the training data (10 questions).

**Evaluation Methods**

In this research, we propose a comprehensive evaluation framework that includes three distinct metrics to assess the machine interpretation of scientific imaging. These metrics encompass overall quality assessment, context assessment, and feature identification assessment, with a particular emphasis on defect and anomaly detection. To ensure a robust and unbiased evaluation, we employ two evaluation tools: GPT-4 and a human evaluator. GPT-4 is used to score the answers generated by our model, with all scores used as is, except for the defect and anomaly assessment. In this case, a human evaluator also intervenes due to specific issues discussed in the defect and anomaly subsection. To account for variability in wording from different papers and language sensitivity in the evaluations, we use the 2023 March-preview of GPT-4 at a temperature and nucleus sampling (top_p) ranges between 0.1 and 0.3. The human evaluator, a Chemical Engineering PhD candidate with a research background in deep learning applied to molecular systems, plays crucial roles in the evaluation process. The human evaluator roles are defining the scoring criteria, providing examples, and engineering and fine-tuning prompts. For clarity, we refer to the answers generated from glass research papers during the data generation step as "benchmark answers," while answers generated by our GlassLLaVA model are referred to as "GlassLLaVA answers."

Overall Quality Assessment

This assessment evaluates the quality of a given response by comparing it with a benchmark answer and the question. We use a 0-100 scale for grading, where a score of 0 indicates a completely inadequate, irrelevant, erroneous, and misleading response, and a score of 100 implies an answer of the same or superior quality to the benchmark. The evaluation begins by reflecting on the benchmark quality, then comparing how well the given response aligns with the following criteria: coverage, accuracy, relevance, fluency, and contextual understanding.

It is worth noting that the submetrics language in this quality assessment could overlap due to the subtle nature of language such as relevance and coverage. Quantitatively, the coverage may be thought of as the area that the GlassLLaVA answer covers relative to the benchmark answer, while the relevance could be measured distance from both answers. While such subtle nuances of language present a challenge here in making distinct evaluations, it also makes the evaluation robust and comprehensive to different qualitative aspects.

Context Assessment

In this evaluation, we extract all simple questions and assess the model's sensitivity to the presence and absence of context that does not directly or indirectly answer the simple question but rather improves the model's understanding of the image by incorporating information such as the materials type, processing conditions, and/or characteristics. We utilize the original question and its corresponding answer to generate three distinct questions, incorporating the same query but with escalating degrees of contextual information (basic, moderate, high). The responses to the newly formulated questions should not be fully included within the questions themselves. The evaluation is performed as in the overall quality assessment.

In our evaluation, we also conducted a detailed examination of the type and quantity of context incorporated into the questions. This analysis was instrumental in understanding the model's sensitivity and response to varying degrees of contextual information. Figs. 2 A, B, and C provide a comprehensive overview of this analysis, illustrating the count of different types of context incorporated into the evaluation questions. The contexts were categorized into five distinct types: material type, appearance, properties, processing conditions, and an 'Other' category for descriptions that did not fit into the other categories. The 'Material Type' context encapsulates information about the specific material being assessed in the SEM image. The 'Appearance' context pertains to the visual attributes or descriptors of the material or image, providing a qualitative assessment of the sample's visual characteristics. The 'Properties' context delves into the inherent or induced characteristics of the material, such as mechanical or thermal properties, offering a deeper understanding of the material's behavior and performance. The 'Processing Conditions' context provides information on the conditions under which the material was processed, including factors like temperature and pressure. Lastly, the 'Other' category accommodates additional context attributes that do not fall into the above categories, potentially encompassing aspects like experimental setup or conditions.

The analysis revealed trends across different levels of context. For the basic context level, the focus was generally on incorporating one additional context attribute per question, often related to the material type or appearance of the sample. As we escalated to moderate context level, we observed the incorporation of two context attributes per question in total. This level tended to bring in more detailed attributes, with a noticeable increase in properties and processing conditions context compared to the basic level. Finally, the high context level included three context attributes per question in total. At this level, there was a substantial rise in the amount of context related to material properties and processing conditions. The 'Other' category also became significantly more prominent as the level of context increased. In terms of general trends, we observed that as the context level escalated from Basic to High, there was a marked increase in the inclusion of material properties, processing conditions, and the 'Other' category. This trend indicates the diversity of attributes provided to the model as the level of context escalates. Material Type and Appearance contexts were consistently present across all levels, but their prominence remained comparatively stable as the level of context escalated. Conversely, Processing Conditions and Properties contexts tended to become more prevalent as more context was added, indicating their crucial role in higher-context questions.

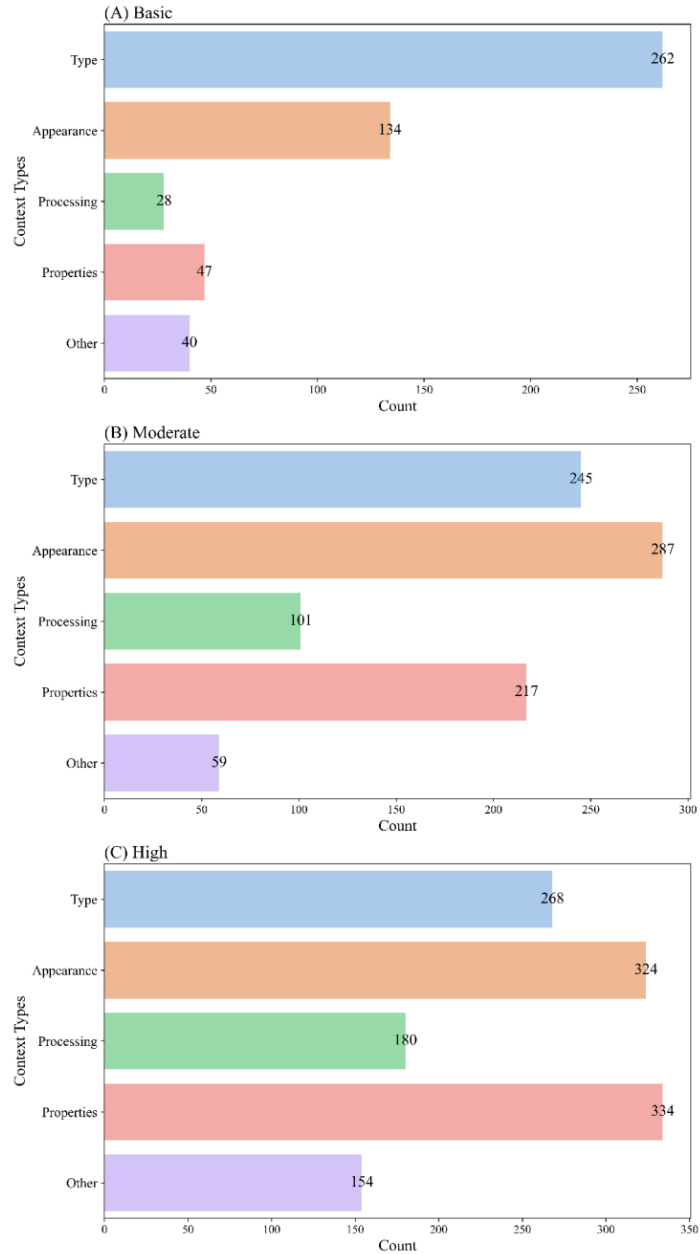

*Figure 2: Count and type of contextual information incorporated into the evaluation questions at three different levels: (A) Basic, (B) Moderate, and (C) High.*

Feature Identification

This assessment evaluates whether GlassLLaVA identifies features consistent with benchmark answers or, framed more on the knowledge scale, whether GlassLLaVA has a general understanding of glass properties and features. The evaluation involves comparing the quality of the microstructure details, for example, in a given response to a provided benchmark answer that involves unseen features. We use a scale from 1-4, where 1.0 means that no relevant features from the benchmark are identified in the response; 2.0 to 3.0 means that very few or minimal similar, relevant, or deducible features from the benchmark are identified in the response; 3.0 to 4.0 means that many similar, relevant, or deducible features from the benchmark are identified in the response; and 4.0 means that most or all similar, relevant, or deducible features from the benchmark are identified, aligning the response closely with the benchmark. We make the score continuous for finer resolution differentiating between few, several, and many features. We also make the assessment under varying levels of context (none, basic, moderate, or high).

Defect and Anomaly Detection

In this evaluation, we aim to determine GlassLLaVA's accuracy in identifying defects and anomalies. The evaluation involves assessing GlassLLaVA's ability to identify or deny the presence of fractures, defects, cracks, or flaws in an image, comparing it with a benchmark answer. Although this assessment can be a subset of feature identification, we assign a binary scale here and we also include the context assessment. That is, we make the assessment under varying levels of context (none, basic, moderate, or high). Furthermore, we only combine true positives/negatives as 1 and false positives/negatives as 0 to avoid complicating the language around this metric. 'NA' is used if the question directly mentions the absence of a fracture/defect or if the benchmark answer indicates that the irrelevance of the detection task. A human evaluator is involved here as well to ensure correct scoring as some language issues seem to be causing NA scores, such as "this may indicate a defect." Furthermore, in some cases, GlassLLaVA indicates that it requires further information to make the judgement; we label such instances as 0 and they tend to be resolved by adding more context.

## Results

We present comprehensive result descriptions of the performance of our multimodal deep learning model, GlassLLaVA, in interpreting scientific imaging. The evaluation is conducted across four distinct assessments: overall quality assessment, context assessment, feature identification, and defect and anomaly detection. Each metric provides a unique perspective on the model's capabilities, offering insights into its strengths and areas for improvement. The results are presented in a systematic manner, with each subsection detailing the findings of a specific metric. The findings are further elucidated through visual representations, such as violin plots and heatmaps, which provide a clear and concise overview of the model's performance.

**Overall Quality Assessment**

The overall quality assessment evaluates the quality of the model's responses by comparing them with benchmark answers and the posed questions. The evaluation employs a 0-100 grading scale, with higher scores indicating superior alignment with the benchmark. Figure 3 presents a violin plot that visualizes the quality of responses across various categories and subcategories.

These results reveal distinct performance trends between simple and complex responses. Generally, simple responses yield lower average score of 67 compared to complex ones, scoring 77 on average. This outcome, although unexpected, is attributed to the presence of more context in the complex questions. Within specific subcategories, the 'Processing and Environmental Impact' subcategory records the highest average score for complex responses of 81, indicating superior alignment with the benchmark. Conversely, the 'Microstructure' subcategory exhibits the lowest average score across both subcategories, with scores of 67 and 77 for simple and complex categories, respectively. The results also reveal variations in quality scores across different categories and subcategories, suggesting diverse quality of responses under different contexts. Despite the observed variability, the model demonstrates strong performance, particularly for complex questions, which are rich in context, and simple questions that lack contextual information.

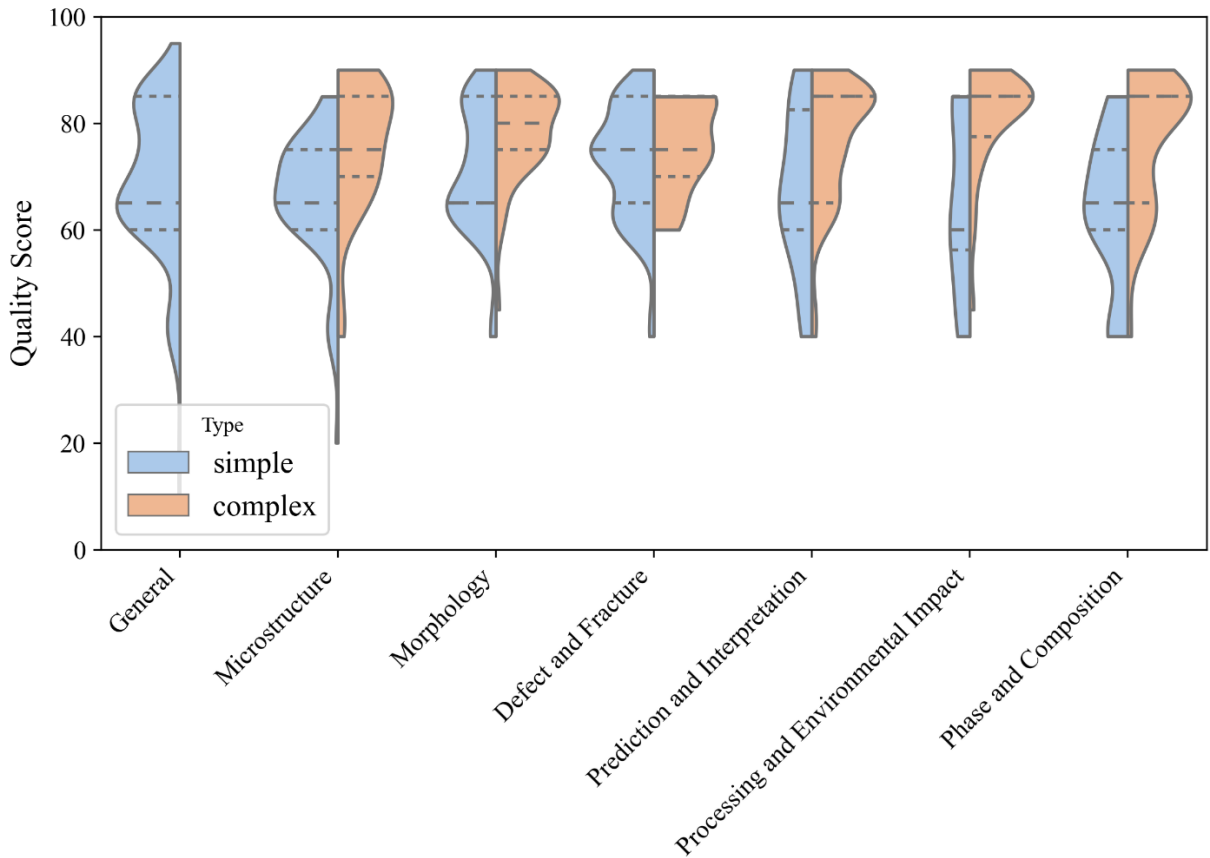

*Figure 3: Violin plot illustrating the distribution of response quality scores across various categories and subcategories (simple and complex), where higher values signify stronger alignment with benchmark answers in quality.*

**Context Assessment**

The context assessment evaluates the model's sensitivity to varying degrees of contextual information. The results, visualized in Figure 4, reveal a consistent trend across all categories: as the degree of context escalates from 'None' to 'High,' the model's scores generally improve. For instance, in the 'General' category, scores rise substantially from 68.84 (None) to 92.56 (High).

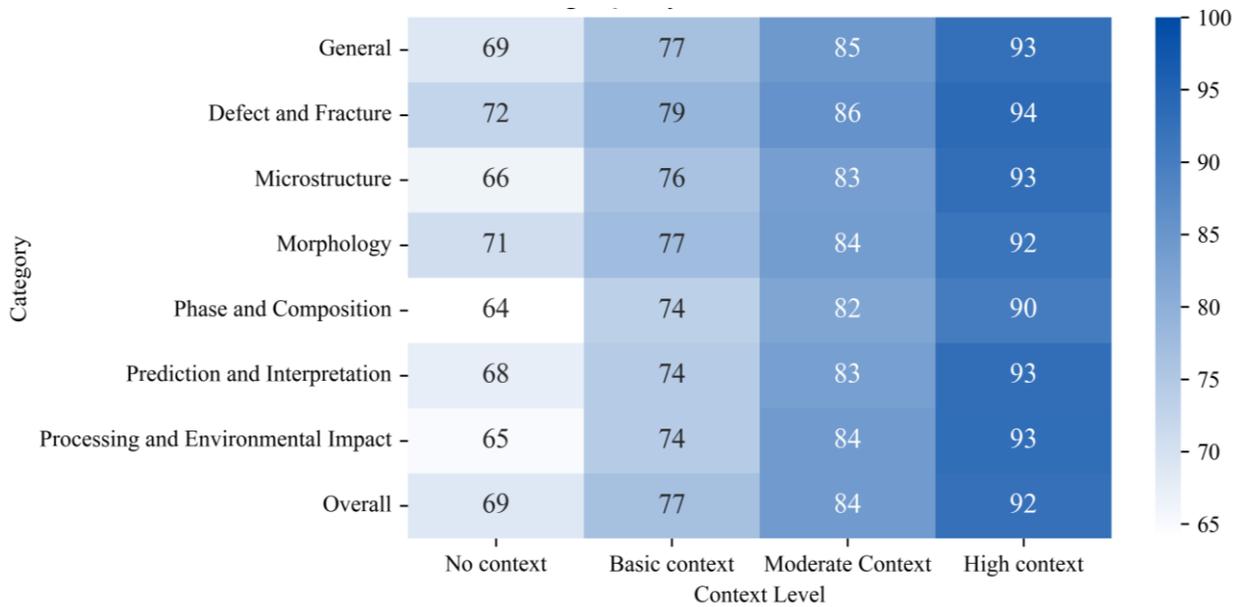

*Figure 4: Heatmap of GlassLLaVA performance across context levels and categories.*

The model demonstrates marked sensitivity to the inclusion of contextual information, as higher context consistently leads to significant increases in scores, suggesting improved model comprehension and response quality. Despite some variability across categories, the presence of more contextual information is consistently correlated with improved assessment scores. This reinforces the value of context in eliciting high-quality, precise responses from the model and offers insights for further optimizing the evaluation process.

**Feature Identification**

The feature identification assessment evaluates whether the model identifies features in responses that are consistent with benchmark answers. The evaluation uses a scale from 1-4, with higher scores indicating closer alignment with the benchmark. The assessment is conducted under varying levels of context: None, Basic, Moderate, High. The results, visualized in Figure 5, reveal that as the level of context increases, the model's scores generally improve. For example, in the 'Morphology' category, scores improve from 2.35 (No context) to 3.58 (High context). This performance may suggest a capacity for generalized understanding of glass properties, encompassing both micro and macro scales.

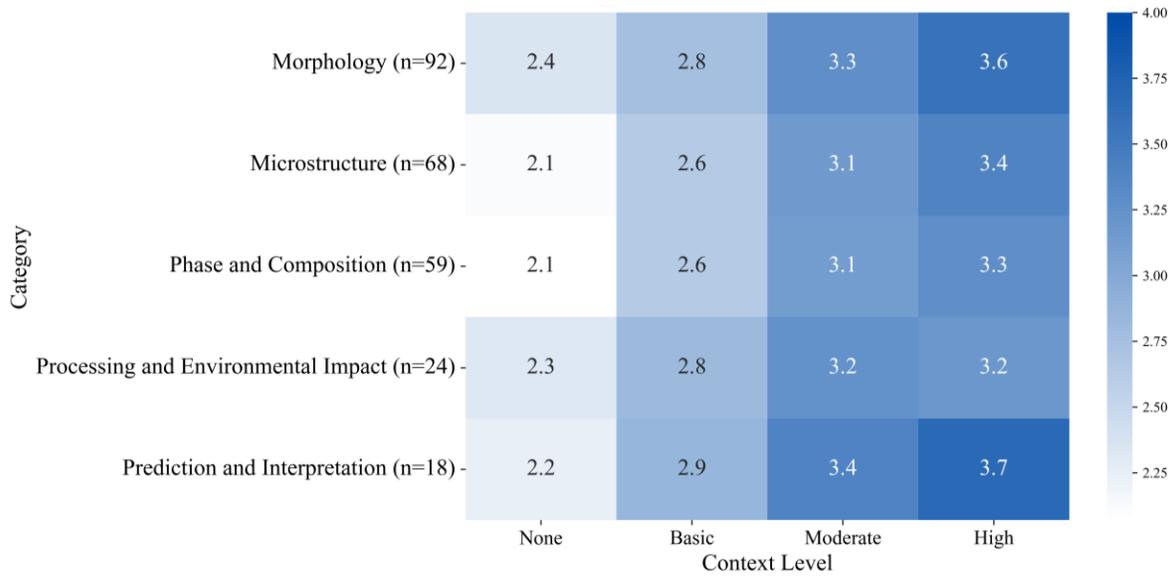

*Figure 5: Heatmap of GlassLLaVA performance on the Feature Identification assessment across context levels and question categories.*

## Defect and Anomaly Detection

The defect and anomaly detection evaluation assesses the model's capability to detect glass defects and abnormalities. The evaluation employs a binary scale, with '1' representing true positives/negatives, and '0' denoting false positives/negatives.

The results, presented in Figure 6, reveal that the model's detection accuracy improves as the level of context increases, from 75% with no context to 95% with high context. Despite not being specifically trained for defect and anomaly detection, GlassLLaVA exhibits a promising ability to identify a variety of glass defects. This suggests that the model has inherently learned relevant features for this task from its training data. However, the data for this analysis largely comes from the glass literature, which predominantly features SEM images with defects. This may introduce a bias in the reported accuracy, as the model may have become more attuned to detecting defects due to their frequent appearance in the data.

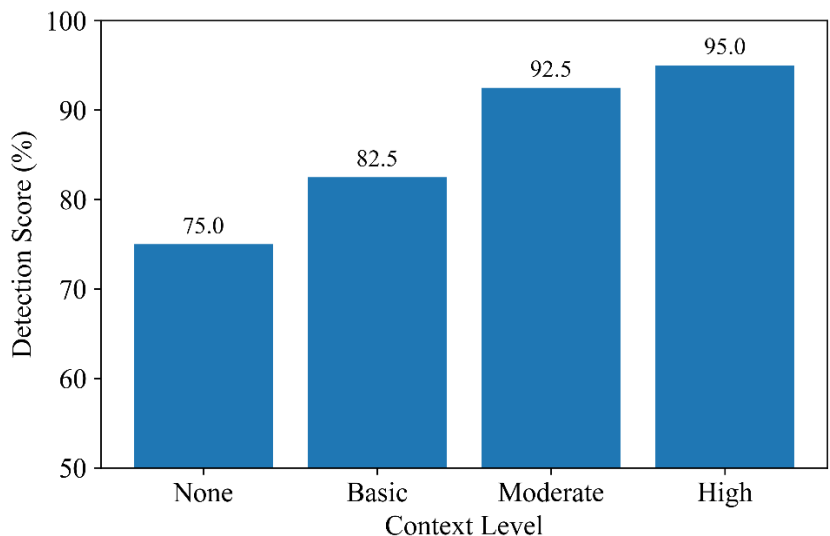

*Figure 6: Defect and anomaly detection accuracy of GlassLLaVA across different context levels.*

## Discussion

Our findings reveal that GlassLLaVA demonstrates a particular proficiency in handling complex questions that are rich in context, as evidenced by higher average scores for complex questions. This performance mirrors human behavior, where experts, equipped with a wealth of contextual knowledge, often excel in interpreting complex scenarios. In this study, GlassLLaVA was trained on 62 papers, aiming to generalize the entire field of glass materials. The model's performance suggests that it has successfully captured the essence of human-like interpretation, particularly in high-context scenarios. Interestingly, the depth and richness of context in complex questions appear to provide the model with more comprehensive data, enabling it to generate high-quality responses. The general trend observed is that the more context our model has, the higher the score it achieves. This suggests that providing as much context as possible to the model can lead to more accurate responses. Furthermore, the performance of the model may be improved if we also provide more context during the training process.

Initially, we aimed to compare our model's performance with human experts and non-experts by having our model's responses rated by both groups. However, this process proved to be time and labor-intensive, particularly when aiming for a large representative sample of glass experts. Considering the potential of extending such an application to a larger dataset or class of materials, we opted to share our work with the community sooner. Furthermore, the benchmark answers extracted from the papers are typically higher than a single human expert response as they are often written and peer-reviewed by multiple experts in the glass field. As such, a survey of experts taken on the individual level may even yield lower scores than the answers reported in papers. Other issues that may be considered are that human expert responses may be shorter, have typos, or lack enough details if taken in a survey, which may even give an inconsistent measure of the actual human performance.

Based on the discussions and results provided above, we ascertain that our model has a close alignment to human expert performance. This is demonstrated by the same human sensitivity to context as well as the results that show similar quality of answers to those presented in peer-reviewed papers. We also assume that our model has a generalized knowledge of glass materials properties as demonstrated by its ability to determine unseen features in the feature identification assessment. As such, GlassLLaVA is a step forward towards bridging the gap between human and machine understanding in scientific imaging interpretation as demonstrated by interpreting SEM images of glass with an accuracy parallel to paper-generated interpretations.

The critical role of the human evaluator could also potentially affect the results and outcome of phrasing criteria to GPT-4. GPT-4 also has a wide range of sensitivities to different parameters and criteria frames. The performance of GPT-4 to provide good questions and answers degrades as the criteria are not framed as bullet points using written language. The temperature and nucleus sampling parameters also affect the grading criteria and the ability of the model to generate unique and fluent questions and answers that are strictly from the papers. The optimal value seems to be around 0.2 for both parameters. Also, for the binary classification in defect and anomaly detection, GPT-4 seems to be sensitive or confused by certain language and the help of a human evaluator that has access to the full paper may be necessary. Also, some of the results are impacted by the data source such as the defect detection where the literature predominantly features defects in glass materials, which might be interesting for further analysis or observing certain patterns. This trend would be different in real-life applications of glass companies where the majority of the glass materials have no defects.

We also explored the correlation between the quality score and the type of context. However, we observed small differences between the combinations of different contexts for the moderate and high context scenarios. The average variations between different combinations are around 2 scores on the 100 scale for the moderate context, as shown in Figure S2, and 1 for the high context. Hence, we conclude that the different combinations are not of significant and are dependent on the question context itself, rather than the context type. This suggests that the influence of context type on model performance is minimal, and the quality of the model's responses is more strongly influenced by the richness and depth of the context provided in the question.

## Conclusion

Our research underscores the transformative potential of LLMs in interpreting scientific imaging, with a specific focus on SEM images of glass materials. The findings suggest that when trained on a rich and diverse dataset, LLMs can generate high-quality responses that closely mirror human expert interpretations. The evaluation metrics we developed for our model, GlassLLaVA, highlight the pivotal role of context in generating accurate interpretations. As the depth and richness of context escalate, the model's performance improves, mirroring the human propensity to excel in interpreting complex scenarios abundant in context. The ability of GlassLLaVA to identify unseen features and detect defects and anomalies in SEM images suggests that the model has developed a generalized understanding of glass materials properties. This implies that LLMs can potentially be trained to comprehend and interpret a wide array of materials, thereby opening up expansive avenues for research and application in the realm of materials science.

The role of the human evaluator in our study underscores the importance of the synergistic collaboration between humans and AI in scientific imaging interpretation. While LLMs can generate high-quality interpretations, human expertise remains crucial in defining the scope and context, providing examples, and fine-tuning the model. Future research could explore ways to enhance this collaboration, potentially by developing more sophisticated evaluation metrics or training larger models to better understand and incorporate human feedback.

Our study also unveils certain limitations and challenges in the application of LLMs in scientific imaging. For instance, the performance of LLMs can be sensitive to the framing of criteria, the richness of context, and the choice of evaluator model parameters. Additionally, the quality of LLM responses can be influenced by the source of the data. These issues underscore the need for further research to optimize the performance of LLMs in scientific imaging interpretation.

Looking ahead, we envision a future where LLMs play a central role in scientific imaging interpretation, aiding researchers in extracting insights from complex images and accelerating scientific discovery. As an assistant to human interpreters, such models could also reduce confirmation bias in scientific imaging by stimulating reevaluation of existing beliefs and hypotheses through LLM linguistic interactions and encouraging a deeper and more profound understanding of materials. To realize such a vision, future research could focus on expanding the application of LLMs to other types of scientific images, wider classes of materials, and larger LLMs. Additionally, developing more sophisticated evaluation metrics and exploring ways to enhance the collaboration between humans and AI could further propel the integration of LLMs in the realm of scientific imaging interpretation.

## Data and Model License

The data, code, and checkpoint files are intended for, and licensed exclusively for, research use only. Usage of these resources must adhere to the license agreements of OpenLLaMA, Vicuna, and GPT-4, as specified under the Apache License 2.0: (Apache License 2.0: [stanford_alpaca/LICENSE at main · tatsu-lab/stanford_alpaca (github.com)](#), [open_llama/LICENSE at main · openlm-research/open_llama (github.com)](#). The datasets used for pretraining are licensed under CC BY NC 4.0 (NonCommercial 4.0 International: [stanford_alpaca/DATA_LICENSE at main · tatsu-lab/stanford_alpaca (github.com)](#), Creative Commons Attribution 4.0 [Creative Commons — Attribution 4.0 International — CC BY 4.0](#)). These licenses permit use for non-commercial purposes only. The model, when trained using these datasets, must not be used for any purpose other than research.

## Acknowledgements

We acknowledge the members of the machine learning team in Corning's Modeling, Software, and Analytics directorate within Manufacturing Technology and Engineering for valuable discussions in the machine learning modeling space, as well as members of the Characterization Sciences group for insight into SEM imaging and providing context about Corning's use of the method within various projects.

# Supporting Information

The following tables and figures are included as supporting information:

**Tables**

- Table S1: Breakdown of training and evaluation sources and images

**Figures**

- Figure S1: K-Means clustering of research papers used in this study

- Figure S2: Variation in quality scores based on context type

**Prompts**:

- Prompt S1 (Q&A Generation): refer to file prompt1.txt

- Prompt S2 (Overall Quality): refer to file prompt2.txt

- Prompt S3 (Context Assessment): refer to file prompt3.txt

- Prompt S4 (Context Type): refer to file prompt4.txt

- Prompt S5 (Feature Identification): refer to file prompt5.txt

- Prompt S6 (Defect and Anomaly): refer to file prompt6.txt

*Table S1: Breakdown by type (training and evaluation) of sources and comma separated names of images in the source papers.*

| Source (doi) | Type | Figures |
| --- | --- | --- |
| 10.1021/jp311138f | Training | Figure 1a, Figure 1b, Figure 1c, Figure 1d |
| 10.1021/nl400304y | Training | Figure 2a, Figure 2b, Figure 2c, Figure 2d, Figure 1b, Figure 1c |
| 10.1021/acsanm.0c00274 | Training | Figure 1a |
| 10.1021/acsphotonics.7b00049 | Training | Figure 2a, Figure 2b, Figure 2c |
| 10.1021/acsami.9b12896 | Training | Figure 1b |
| 10.1021/mz500725s | Training | Figure 2a, Figure 2b, Figure 2c, Figure 2d, Figure 2e, Figure 2f |
| 10.1021/am401953j | Training | Figure 2b, Figure 3, Figure 4, Figure 5, Figure 6, Figure 8 |
| 10.1021/acs.cgd.5b00253 | Training | Figure 4, Figure 4, Figure 4, Figure 4 |
| 10.1021/am100746c | Training | Figure 2 |
| 10.1021/cg3002702 | Training | Figure 2 |
| 10.1021/cg3009909 | Training | Figure 4 |
| 10.1021/la502873d | Training | Figure 1a, Figure 1b |
| 10.1021/acssuschemeng.0c04026 | Training | Figure 2a, Figure 2b, Figure 2c, Figure 2d |
| 10.1021/acsaem.1c01645 | Training | Figure 1d, Figure 2d |
| 10.1021/acsami.0c17897 | Training | Figure 1c, Figure 1f |
| 10.1021/acsomega.0c01307 | Training | Figure 4a, Figure 4b, Figure 4c |
| 10.1021/acsaem.1c00402 | Training | Figure 1b, Figure 1c, Figure 1d |
| 10.1021/acsami.8b02338 | Training | Figure 2a, Figure 2b, Figure 2c, Figure 2d, Figure 2e, Figure 4a, Figure 4b, Figure 4c, Figure 4d, Figure 4e, Figure 5a, Figure 5b, Figure 5c, Figure 5d, Figure 5e |
| 10.1021/acsami.3c02040 | Training | Figure 5 d, Figure 5 e, Figure 5 g, Figure 5 h |
| 10.1021/acsami.2c18774 | Training | Figure 2 a, Figure 2 b, Figure 2 c, Figure 2 d, Figure 2 e |
| 10.1021/cg4008087 | Training | Figure 2 annealed at 790 °C, Figure 2 annealed at 810 °C, Figure 2 annealed at 830 °C, Figure 2 annealed at 850 °C, Figure 2 annealed at 970 °C, Figure 5 a, Figure 5 b, Figure 5 c, Figure 5 d, Figure 7a, Figure 7b, Figure 7c, Figure 7d, Figure 9 |
| 10.1021/acsomega.2c00365 | Training | Figure 4a, Figure 4b, Figure 5a, Figure 5b, Figure 6a, Figure 6b, Figure 7a, Figure 7b, Figure 8a, Figure 8b, Figure 8c, Figure 8d |
| 10.1021/jp9091078 | Training | Figure 1a, Figure 1b, Figure 1c |
| 10.1016/j.nocx.2022.100124 | Training | Fig. 1a, Fig. 1b, Fig. 1c |
| 10.1016/j.ssi.2015.09.021 | Training | Fig. 4a, Fig. 4b, Fig. 4c, Fig. 5a, Fig. 5b |
| 10.1016/j.prosdent.2022.09.014 | Training | Figure 5A, Figure 5B, Figure 5C, Figure 5D, Figure 5E, Figure 5F, Figure 5G, Figure 5H, Figure 5I, Figure 5J, Figure 2A, Figure 2B, Figure 2C, Figure 2D, Figure 2E, Figure 2F |
| 10.1016/j.jnoncrysol.2022.121940 | Training | Fig. 6(a), Fig. 6(b), Fig. 6(c), Fig. 6(d), Fig. 12(a), Fig. 12(a) |
| 10.1016/j.jnoncrysol.2022.122113 | Training | Fig. 7 A4, Fig. 7 A5, Fig. 7 A6, Fig. 7 A7 |
| 10.1016/j.jnoncrysol.2022.121436 | Training | Fig. 4 1h, Fig. 4 1.5h, Fig. 4 2h, Fig. 4 2.5h |
| 10.1016/j.ceramint.2022.02.105 | Training | Fig. 6. GC before glass formation process, Fig. 6. GC after glass formation process, Fig. 6. RGC1 before glass formation process, Fig. 6. RGC1 after glass formation process, Fig. 6. RGC2 before glass formation process, Fig. 6. RGC2 after glass formation process, Fig. 6. RGC3 before glass |
| | | formation process, Fig. 6. RGC3 after glass formation process, Fig. 6. RGC4 before glass formation process, Fig. 6. RGC5 after glass formation process |
| 10.1016/j.jnucmat.2022.153879 | Training | Fig. 4 (a) Gd3, Fig. 4 (b) Gd5, Fig. 4 (c) Gd7, Fig. 4 (d) Gd10, Fig. 4 (e) Gd15 |
| 10.1016/j.ssi.2022.116106 | Training | Fig. 7 (a), Fig. 7 (b), Fig. 7 (c), Fig. 7 (c), Fig. 7 (d), Fig. 7 (d), Fig. 7 (e), Fig. 7 (f) |
| 10.1016/j.ceramint.2023.06.048 | Training | Fig. 3 a) S-TV, Fig. 3 b) S-Tv/Ti10,, Fig. 3 c) S-TV/Ti20. |
| 10.1016/j.surfcoat.2021.127495 | Training | Fig. 5a, Fig. 5b, Fig. 11. (a) without SiO2 interlayer for 1 h, Fig. 11. (b) SiO2 interlayer for 1 h, Fig. 11. (c) without SiO2 interlayer for 10h, Fig. 11. (d) SiO2 interlayer for 10h |
| 10.1016/j.jnoncrysol.2021.120810 | Training | Fig. 3b. G10E, Fig. 3b. G20E, Fig. 3b. G30E, Fig. 3b. G40E, Fig. 3b. G50E, Fig. 3b. G60E, Fig. 3b. G70E |
| 10.1016/j.materresbull.2018.04.007 | Training | Fig. 3. GC1, Fig. 3. GC2, Fig. 3. GC3, Fig. 3. GC4 |
| 10.1016/j.optmat.2021.111772 | Training | Fig. 4(A) 690 °C/1h, Fig. 4(A) 700 °C/1h, Fig. 4(A) 710 °C/1h, Fig. 4(A) 720 °C/1h, Fig. 4(B) |
| | | 710 °C/1h, Fig. 4(B)  710 °C/1.5h, Fig. 4(B)  710 °C/2h, Fig. 4(B)  710 °C/2.5h |
| 10.1016/j.ceramint.2021.04.108 | Training | Fig. 6. a, Fig. 6. b, Fig. 6. c, Fig. 6. a, Fig. 6. b, Fig. 6. c, Fig. 7. a, Fig. 7. b, Fig. 7. c, Fig. 7. a, Fig. 7. b, Fig. 7. c, Fig. 8. a, Fig. 8. b, Fig. 8. c, Fig. 8. a, Fig. 8. b, Fig. 8. c |
| 10.1016/j.jmrt.2021.07.158 | Training | Fig.4 (a) S1, Fig.4 (b) S2, Fig.4 (c) S3, Fig.4 (d) S4, Fig.4 (e) S5, Fig.4 (f) S4 |
| 10.1016/j.matpr.2021.05.184 | Training | Fig. 5, Fig. 6, Fig. 7, Fig. 8 |
| 10.1016/j.matpr.2020.02.852 | Training | Fig. 3(a), Fig. 3(b), Fig. 4(a), Fig. 4(b) |
| 10.1016/j.istruc.2022.02.053 | Training | Fig. 20a, Fig. 20a, Fig. 20a, Fig. 20b, Fig. 20b, Fig. 20b, Fig. 21a, Fig. 21b, Fig. 21c, Fig. 21d, Fig. 21e, Fig. 21f, Fig. 22a, Fig. 22b, Fig. 22c, Fig. 22d, Fig. 22e, Fig. 22f |
| 10.1016/j.jclepro.2022.134576 | Training | Fig. 9 M1.4N6-46, Fig. 9 M1.4N10-46, Fig. 10 M1.2N8-46, Fig. 10 M1.6N8-46, Fig. 10 M1.8N8-46, Fig. 11. 1d, Fig. 11. 7d, Fig. 11. 28d |
| 10.1016/j.jmapro.2020.08.068 | Training | Fig. 9a, Fig. 9b, Fig. 9c, Fig. 9d, Fig. 10a, Fig. 10b, Fig. 10c, Fig. 10d, Fig. 10e, Fig. 10f, Fig. 11a, Fig. 11b, Fig. 11c, Fig. 11d, Fig. 11e |
| 10.1016/j.jnoncrysol.2022.121530 | Training | Fig. 3a, Fig. 3b, Fig. 3c, Fig. 3d, Fig. 3e, Fig. 5a, Fig. 5b, Fig. 5c, Fig. 5d, Fig. 7a, Fig. 7b, Fig. 7c, Fig. 7d |
| 10.1016/j.jnoncrysol.2022.121644 | Training | Fig. 5a, Fig. 5b, Fig. 5c |
| 10.1016/j.ceramint.2022.08.124 | Training | Fig. 1 GC15, Fig. 1 GC20, Fig. 1 GC25, Fig. 1 GC30 |
| 10.1016/j.ceramint.2017.10.079 | Training | Fig. 3a, Fig. 3b, Fig. 3c, Fig. 3d |
| 10.1016/j.ceramint.2022.06.166 | Training | Fig. 4 a, Fig. 4 b, Fig. 4 c, Fig. 6 a, Fig. 6 b, Fig. 6 c, Fig. 8 a, Fig. 8 b, Fig. 9 a, Fig. 9 b, Fig. 9 c, Fig. 9 d |
| 10.1016/j.mtcomm.2017.02.007 | Training | Fig. 5 a, Fig. 5 b, Fig. 5 c, Fig. 5 d, Fig. 6 a, Fig. 6 b, Fig. 6 c, Fig. 7 |
| 10.1016/j.jeurceramsoc.2021.05.018 | Training | Fig. 10 a, Fig. 10 c, Fig. 10 e, Fig. 10 b, Fig. 10 d, Fig. 10 f, Fig. 12 |
| 10.1016/j.jallcom.2019.06.107 | Training | Fig. 3a, Fig. 3b, Fig. 3c, Fig. 3d, Fig. 4a, Fig. 4b, Fig. 4c, Fig. 4d, Fig. 5a, Fig. 5a |
| 10.1016/j.ceramint.2018.05.054 | Training | Fig. 1A, Fig. 1B, Fig. 1C |
| 10.1016/j.engfailanal.2018.04.007 | Training | Fig. 6, Fig. 6, Fig. 7b fracture surface, Fig. 7b cross section, Fig. 7b fracture surface, Fig. 7b cross section |
| 10.1016/j.jnoncrysol.2010.07.068 | Training | Fig. 5a Bonded borate glass scaffold, Fig. 5a Bonded borate glass scaffold, Fig. 5b scaffold immersed for 1 day in K2HPO4 solution, Fig. 5b scaffold immersed for 1 day in K2HPO4 solution, Fig. 5b scaffold immersed for 4 day in K2HPO4 solution, Fig. 5b scaffold immersed for 4 day in K2HPO4 solution, Fig. 5b scaffold immersed for 7 day in K2HPO4 solution, Fig. 5b scaffold immersed for 7 day in K2HPO4 solution |
| 10.1016/j.conbuildmat.2020.119971 | Training | Fig. 2a, Fig. 2b, Fig. 2c, Fig. 2d, Fig. 2e, Fig. 2f, Fig. 3a, Fig. 3b, Fig. 3c, Fig. 3d, Fig. 3e, Fig. 3f |
| 10.1016/j.tsf.2017.09.016 | Training | Fig. 3e , Fig. 3f, Fig. 5a, Fig. 5b, Fig. 5c, Fig. 5d, Fig. 5e, Fig. 5f, Fig. 7a, Fig. 7b, Fig. 7c, Fig. 7d |
| 10.1016/j.jeurceramsoc.2012.12.012 | Training | Fig. 6 a, Fig. 6 b, Fig. 6 c |
| 10.1016/j.jeurceramsoc.2016.10.020 | Training | Fig. 6a, Fig. 6e, Fig. 6b, Fig. 6f, Fig. 6c, Fig. 6g, Fig. 6d, Fig. 6h, Fig. 7a, Fig. 7e, Fig. 7b, Fig. 7f, Fig. 7c, Fig. 7g, Fig. 7d, Fig. 7h |
| 10.1016/j.jnoncrysol.2015.03.022 | Training | Fig. 5a, Fig. 5b, Fig. 5c, Fig. 5d, Fig. 5e, Fig. 5f, Fig. 13 G1 1 hour, Fig. 13 G1 12 hours, Fig. 13 G1 24 hours, Fig. 13 G2 1 hour, Fig. 13 G2 12 hours, Fig. 13 G2 24 hours, Fig. 13 G3 1 hours, Fig. 13 G3 12 hour, Fig. 13 G3 24 hours, Fig. 13 G4 1 hour, Fig. 13 G4 12 hours, Fig. 13 G4 24 hours |
| 10.1007/s10853-018-2284-6 | Evaluation | Fig. 2, Fig. 2 I A and B, Fig. 2 II C, Fig. 3a, Fig. 3a I, Fig. 3a II, Fig. 3b, Fig. 3b I, Fig. 3b II, Fig. 3c, Fig. 3c I, Fig. 3c II, Fig. 3d, Fig. 3d I, Fig. 3d II, Fig. 4a, Fig. 4a I, Fig. 4a II, Fig. 4b, Fig. 4b I, Fig. 4b II, Fig. 4c, Fig. 4c I, Fig. 4c II, Fig. 4d, Fig. 4d I, Fig. 4d II, Fig. 4e, Fig. 4e I, Fig. 4e II |
| 10.1007/s11041-016-0041-5 | Evaluation | Figure 1 c, Fig. 4 a, Fig. 4 b, Fig. 4 c, Fig. 4 d, Fig. 4 e, Fig. 5, Figure 6 |
| 10.1038/s41598-019-44384-z | Evaluation | Fig. 2b 250 nm thick, Fig. 2c 60 nm thick, Fig. 2d 16 nm thick, Fig. 2f 9 nm thick, Fig. 2e 7 nm thick |
| 10.1134/S1087659622601083 | Evaluation | Fig. 1a, Fig. 1b |
| 10.1134/S1027451017040218 | Evaluation | Figure 1a, Figure 1b, Figure 2, Figure 3, Figure 4 |
| 10.1007/s12206-020-1017-3 | Evaluation | Fig.3 1100C 500 um, Fig.3 1100C 50 um, Fig.3 1200C 500 um, Fig.3 1200C 50 um, Fig.3 1225C 500 um, Fig.3 1225C 50 um, Fig.3 1250C 500 um, Fig.3 1250C 50 um, Fig.3 1300C 500 um, Fig.3 1300C 50 um |
| 10.1134/S1087659617060049 | Evaluation | Figure 5 sample C1, Figure 5 sample C2, Figure 5 sample D0, Figure 5 sample D1, Figure 5 sample D2 |
| 10.1134/S0040579516020020 | Evaluation | Figure 3 a 750°C, Figure 3 b 850°C, Figure 3 c 950°C, Figure 3 d 1050°C, Figure 3 e 1250°C |
| 10.1134/S1087659617020110 | Evaluation | Fig. 1a sample 23, Fig. 1a sample 11, Fig. 2a sample 2, Fig. 2b sample 6 20um, Fig. 2c sample 6 2um |
| 10.1134/S0020168520100064 | Evaluation | Figure 4, Figure 5 |

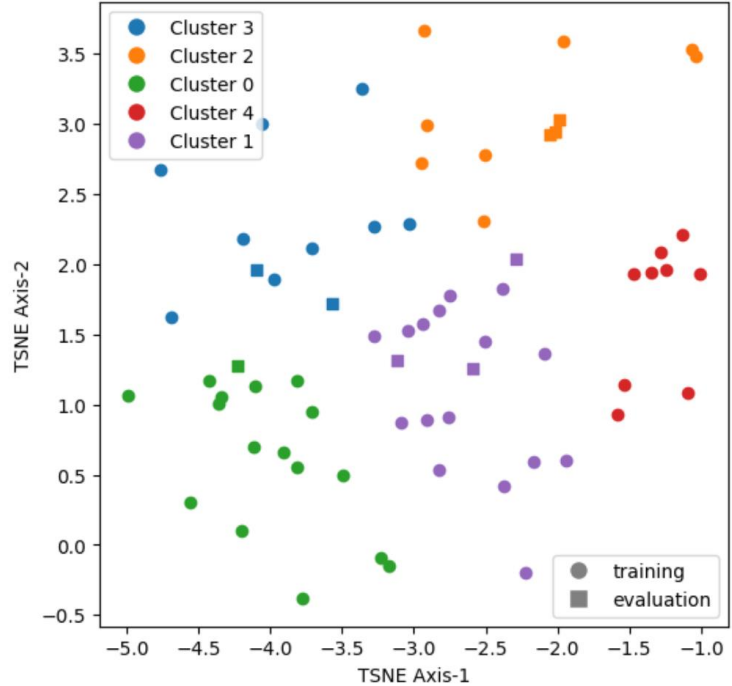

*Figure S1: K-Means clustering of research papers used in this study*

In this clustering analysis (Figure S1), document embeddings are first generated for a collection of research papers using a specified text embedding engine "text-embedding-ada-002" from OpenAI. These embeddings are then reduced to a two-dimensional space using the t-SNE (t-Distributed Stochastic Neighbor Embedding) algorithm. After this dimensionality reduction, the K-Means algorithm is applied to cluster the papers into five distinct groups. The papers are labeled either as 'training' or 'evaluation'. The resultant clusters, along with the type of each paper, are visualized in a scatter plot. Different colors are used to represent different clusters, and distinct markers are used to indicate whether a paper is part of the 'training' or 'evaluation' set. The clusters are assigned titles by GPT-4 as follows:

- **Cluster 0**: Advancements in Glass and Glass-Ceramic Materials: Applications in Lighting, Photovoltaics, Nanotechnology, and Tissue Engineering
- **Cluster 1**: Advancements in Material Science: Investigating the Effects of Nanoparticle Incorporation and Environmental Factors on the Mechanical and Morphological Properties of Fiber-Reinforced Polymer Composites and Metallic Glasses

- **Cluster 2**: Advancements in the Fabrication, Characterization, and Applications of Glass-Ceramics: A Focus on Electrochemical Performance, Structural Stability, and Material Optimization
- **Cluster 3**: "Crystallization Mechanisms, Synthesis, and Luminescent Properties of Doped Glass-Ceramics for Advanced Applications
- **Cluster 4**: Advancements in Glass Manipulation and Modification: Nanocapillaries, Microlens Arrays, and Laser Processing Techniques

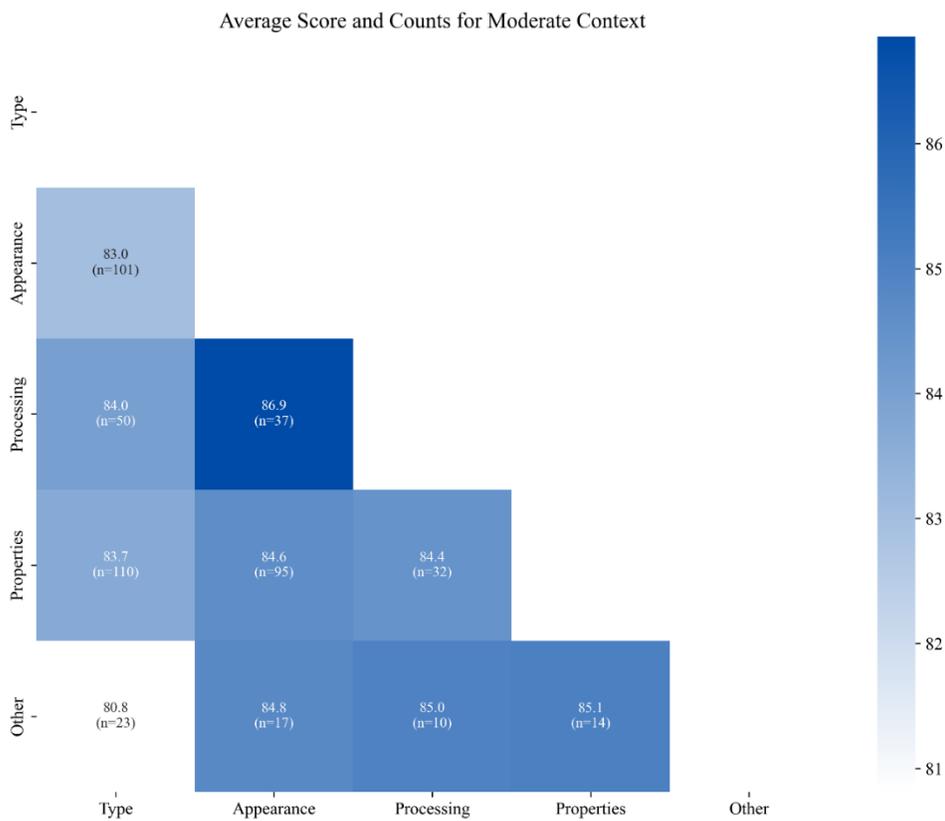

*Figure S2: Variation in quality scores based on context type*

Figure S2 illustrates the minor differences in quality scores across various context combinations, highlighting that the model's response quality is more significantly influenced by the richness and depth of the question's context, rather than the type of context itself.